\documentclass[11pt]{article}
\usepackage[utf8]{inputenc}
\usepackage[T1]{fontenc}

\usepackage[margin=1in]{geometry}

\usepackage{lmodern}
\usepackage{microtype}

\usepackage{graphicx}
\usepackage{amsmath,amssymb}
\usepackage{algorithm}
\usepackage{algpseudocode}
\usepackage{caption}
\usepackage{booktabs}
\usepackage{tabularx}
\usepackage{multirow}
\usepackage{threeparttable}
\usepackage{colortbl}
\usepackage{xspace}              

\newcommand{\etal}{\emph{et al.}\xspace}

\usepackage[colorlinks,allcolors=blue]{hyperref}

\usepackage[
  backend=biber,       
  style=ieee,          
  sorting=nty,         
  maxnames=3,          
]{biblatex}
\usepackage[margin=1in]{geometry}
\usepackage{tikz}
\usepackage{graphicx}
\usepackage{lmodern}

\begin{document}

\begin{center}
  \rule{\textwidth}{1pt}\par
  \vspace{1em}

  {\LARGE\scshape
    Pairwise Alignment \& Compatibility for Arbitrarily Irregular Image Fragments
  }\par
  \rule{\textwidth}{1pt}
\end{center}

\vspace{1em}

\noindent
\begin{minipage}{0.32\textwidth}\centering
  \textbf{Ofir Itzhak Shahar}\\
  Dept.\ of Computer Science\\
  Ben-Gurion University of the Negev\\
  \texttt{shofir@post.bgu.ac.il}
\end{minipage}\hfill
\begin{minipage}{0.32\textwidth}\centering
  \textbf{Gur Elkin}\\
  Dept.\ of Computer Science\\
  Ben-Gurion University of the Negev\\
  \texttt{gurshal@post.bgu.ac.il}
\end{minipage}\hfill
\begin{minipage}{0.32\textwidth}\centering
  \textbf{Ohad Ben-Shahar}\\
  Dept.\ of Computer Science\\
  Ben-Gurion University of the Negev\\
  \texttt{ben-shahar@cs.bgu.ac.il}
\end{minipage}

\vspace{3em}

\begin{center}
  \textbf{ABSTRACT}
\end{center}


Pairwise compatibility calculation is at the core of most fragments-reconstruction algorithms, in particular those designed to solve different types of the jigsaw puzzle problem. However, most existing approaches fail, or aren’t designed to deal with fragments of realistic geometric properties one encounters in real-life puzzles. And in all other cases, compatibility methods rely strongly on the restricted shapes of the fragments. In this paper, we propose an efficient hybrid (geometric and pictorial) approach for computing the optimal alignment for pairs of fragments, without any assumptions about their shapes, dimensions, or pictorial content. We introduce a new image fragments dataset generated via a novel method for image fragmentation and a formal erosion model that mimics real-world archaeological erosion, along with evaluation metrics for the compatibility task. We then embed our proposed compatibility into an archaeological puzzle-solving framework and demonstrate state-of-the-art neighborhood-level precision and recall on the RePAIR 2D dataset, directly reflecting compatibility performance improvements.


\section{Introduction \& related work}
\label{sec:intro}

Although computer vision and image processing have been experiencing significant breakthroughs in the last decade, some real-world problems, even those that could be abstracted and formulated relatively easily, remain unsolved. Such is the problem of robustly reconstructing real-world archaeological artifacts from their fragments, a challenge that is a major motivation for seeking computational solutions to the \textit{jigsaw puzzle problem} (e.g.,~\cite{pomeranz2011, sağıroğlu2006texture,  sağıroğlu2010optimization, gur2017square, derech2021solving, vardi2023multi, Khoroshiltseva2024, Harel2024}) and is still largely open despite the vast resources, time, and direct involvement of domain experts recruited to its solution~\cite{Caple2020, Eslami2020, RePAIR2021, brown2008system}.

Freeman and Garder first posed the 2D jigsaw-puzzle problem for unrestricted shapes in 1964 \cite{freeman1964}, later shown to be NP-complete \cite{demaine2007jigsaw}. Since then most research has focused on square-piece puzzles—now even addressed by deep learning \cite{Song2023, kim2025, talon2025}—while the original case of arbitrary-shape fragments remains unsolved. To date, to our best knowledge, no solver successfully reconstructs public archaeological datasets without relying on prior knowledge of the target image \cite{Tsesmelis2024, Dondi2020DAFNE, barra2020, lerme2020, daSilvaTeixeira2021}.

\begin{figure}[]
    \centering
    \begin{tabular}{ccc}
        \includegraphics[width=0.3\columnwidth]{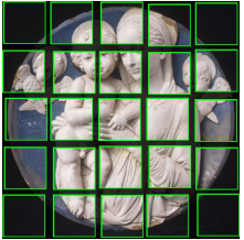} 
        &
        \includegraphics[width=0.3\columnwidth]{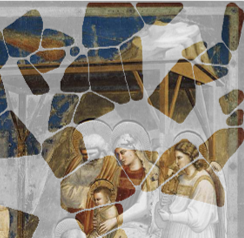}
        &
        \includegraphics[width=0.3\columnwidth]{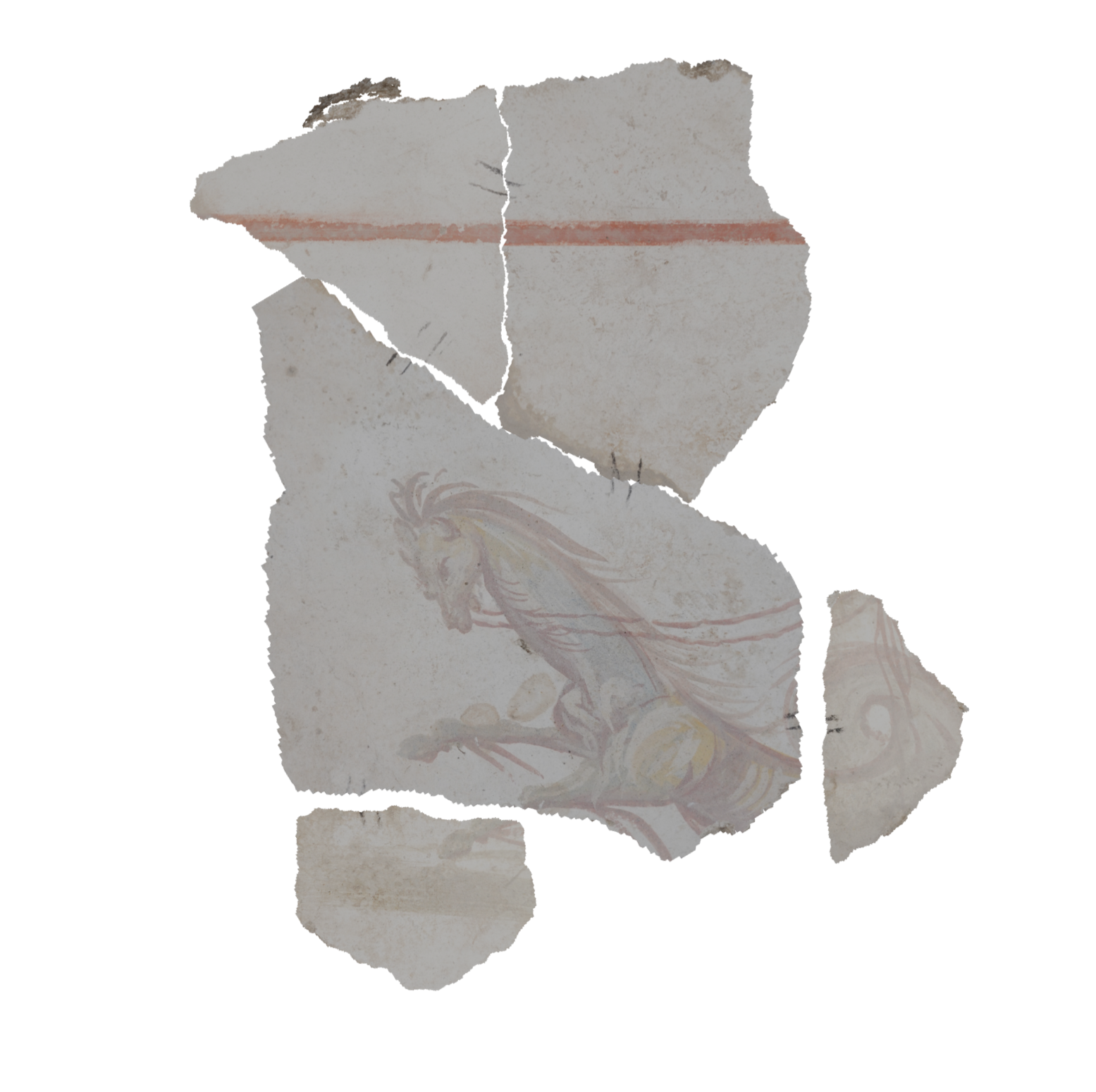}
        \\
        A & B & C 
        \end{tabular}
    \caption{
    Samples from publicly available puzzle datasets.
    \textbf{A:}~A Square Jigsaw puzzle from JPwLEG-5 dataset \cite{Song2023}. published in 2023 along with JPwLEG-3, both containing strictly square puzzles with 25 / 9 fragments respectively, while containing gaps between neighboring fragments.    
    \textbf{B:}~A region from a fragmented puzzle of the fresco \textit{Giotto, Adoration of the Magi}, sampled from the DAFNE dataset~\cite{Dondi2020DAFNE}. The fragments are created by a Voronoi partition with a smoothing-based erosion, leaving the eroded pieces with relatively simple shapes.
    \textbf{C:}~An archaeological puzzle, sampled from the RePAIR dataset \cite{Tsesmelis2024}, which is based on 3D scanned fragments of a broken fresco from Pompeii, and 2D rendered versions of their pictorial \textit{surface}. All fragments exhibit real-world erosion and unrestricted complex shapes, with varying degrees of pictorial content. 
    }
    \label{fig:puzzles_from_recent_datasets}
\end{figure}

An additional important aspect of the archaeological reconstruction domain, which is often treated poorly or completely ignored by recent works, is the potential  \textit{erosion} on the archaeological fragments (see Fig.~\ref{fig:puzzles_from_recent_datasets}).
In realistic scenarios, fragments may be degraded due to poor handling, harsh physical conditions, or simply matter deterioration over time which might drastically alter their physical geometric shape, as is often observed in real pieces \cite{Tsesmelis2024}. Unfortunately, even when this is taken into consideration, recent works on puzzle solving and archaeological reconstruction tend to simulate it in ways that rarely resemble realistic scenarios~\cite{Yang2024, bridger2020solving, rika2025generic, chen2023, Song2023}. 
%

It should be mentioned that unlike in the computer-vision literature, efforts to handle puzzles of unrestricted geometric shapes were made in the computational archaeology literature, often while ignoring pictorial content  \cite{brown2008system, Castaeda2011GlobalCI, TolerFranklin2010MultifeatureMO, Funkhouser2011LearningHT, Brown2012ToolsFV, pintus2014geometric, Pintus2016ASO}. These works typically focus on 3D rather than 2D reconstruction and tend to follow a similar computational pipeline. In this pipeline, the fragments are scanned into point clouds, processed into meshes, and segmented into facets, while geometrical features, and occasionally simple pictorial features (such as average color, saturation, and variance) are extracted either from these facets or their boundary curves. Candidate pairwise matches of fragments are then computed utilizing these features, while often optimizing geometrical coherency (e.g., \cite{brown2008system, Castaeda2011GlobalCI}). 



As implied above, only few existing approaches are even capable of running on (and let alone solving) general problems at the complexity of \textit{archaeological} puzzles with no limitations on the fragment shapes, sizes, and content. 
Among these works, 
Derech \etal~\cite{derech2021solving} proposed a greedy reconstruction algorithm that iteratively picks the best additional fragment to add to a growing collection of reconstructed pieces, discretizing the infinite space of orientations into a coarse finite set and scoring solely pictorial compatibility via pixel‐wise dissimilarity on extrapolated content. However, relying on a regular, equally-spaced discretization of the configuration space creates a trade-off between runtime and the ability to capture optimal alignments, potentially missing better matches due to coarse sampling. Additionally, their outdated extrapolation prohibits useful results on real archaeological data. 
Tsesmelis \etal~\cite{Tsesmelis2024} present two geometric schemes—one replacing pictorial matching with Harel \etal’s spring‐mass optimization~\cite{Harel2024}, the other a genetic global optimizer balancing reconstruction area and overlap—both ignoring pictorial coherence. Cao \etal~\cite{Cao2024} employ a two‐stage global reassembly with region‐overlap pruning and a MobileViT classifier to score polygonal‐edge alignments but do not model erosion. Khoroshiltseva \etal~\cite{Khoroshiltseva2024} handle irregular shapes via line‐continuation features and limited orientations, while Puzzlefusion~\cite{hosseini2023puzzlefusion} and the Crossing Cut solver~\cite{Harel2024} depend on low‐degree polygonal approximations ill‐suited to unrestricted, eroded fragments. Overall, these methods impose substantial constraints on shape, orientation, or compatibility scoring and neglect the combined challenges of unrestricted geometries, pictorial content, and realistic erosion.

In this work we propose a different approach that does not restrict fragment shapes, sizes, pictorial content, or the potential loss of information (both geometric and pictorial) due to erosion. 
We deliberately choose to focus on what might be considered the most important part of any puzzle-solving algorithm, namely the determination of compatibility and relative alignment between potentially neighboring pairs of pieces. Toward that goal, we also propose an \textit{adaptive} discretization method to constrain the infinite space of potential relative configurations between fragments to a tractable size, in an informed way based on their unique shapes. 
We thus propose \textbf{PolEx}, a \textit{Polygonal approximation \& Extrapolated Pictorial dissimilarity}, whose outline include the following steps:

\begin{itemize}
    \item \textbf{1 - Extrapolation} of the pictorial content of all fragments, while utilizing an advanced diffusion model~\cite{Rombach2022}, and the extraction of \textit{extrapolated bands}.
    
    \item \textbf{2 - Rich multiscale polygonal approximation} of the fragments' original geometric shapes, including \textit{augmented} edges representing larger scale geometric segments of the boundary.
    
    \item \textbf{3 - Extraction of Candidate Matching Edges} between the rich polygonal approximations of the two fragments under consideration. 
    
    \item \textbf{4 - Computing Potential Alignment Configurations} for the inspected edges.
    
    \item \textbf{5 - Scoring potential configurations} based on pictorial dissimilarity of random patches sampled across the shared region of the fragments' extrapolated bands.
\end{itemize}

\section{Archaeologically-inspired fragmentation}
\label{sec:data}

Although many have addressed puzzle-solving challenges, only few archaeological reconstruction datasets are currently available, and none are specifically designed to test pairwise pictorial dissimilarity and alignment. To address this gap and asses our pairwise alignment and compatibility approach, we present a novel dataset of fragmented images that better simulate the geometry of broken archaeological artifacts. 
The pieces are generated using a controlled fragmentation process, inspired by contemporary reconstruction challenges~\cite{Dondi2020DAFNE,Tsesmelis2024}. The realism of our synthetic approach is most strongly apparent in the \textit{uneven degree of erosion} along fragment boundaries, thus incorporating another aspect of complexity for reassembly algorithms. 

\subsection{Image generation}

In total, our dataset comprises 1000 archaeological puzzles. To generate the base images, we first employed a LLM \cite{brown2020} to create approximately 1400 specialized prompts, seeking to describe high-quality fresco images across various artistic styles~\cite{elkin2025recognizing}. These prompts were then fed into Stable-Diffusion 3.5~\cite{Rombach2022} to generate the diverse set of images used in our dataset.

Our decision to use synthetically generated images rather than photographs of actual frescos serves multiple purposes: 
(1) it ensures no publicly available images can be exploited in the reconstruction process (for example, by granting the solver internet access for reference materials); 
(2) it provides precise control over image properties that affect reconstruction difficulty, such as detail size and color variance; and 
(3) it avoids potential copyright and cultural heritage considerations while still producing convincing fresco-like images.



\begin{figure*}
    \centering
    \includegraphics[width=0.9\textwidth]{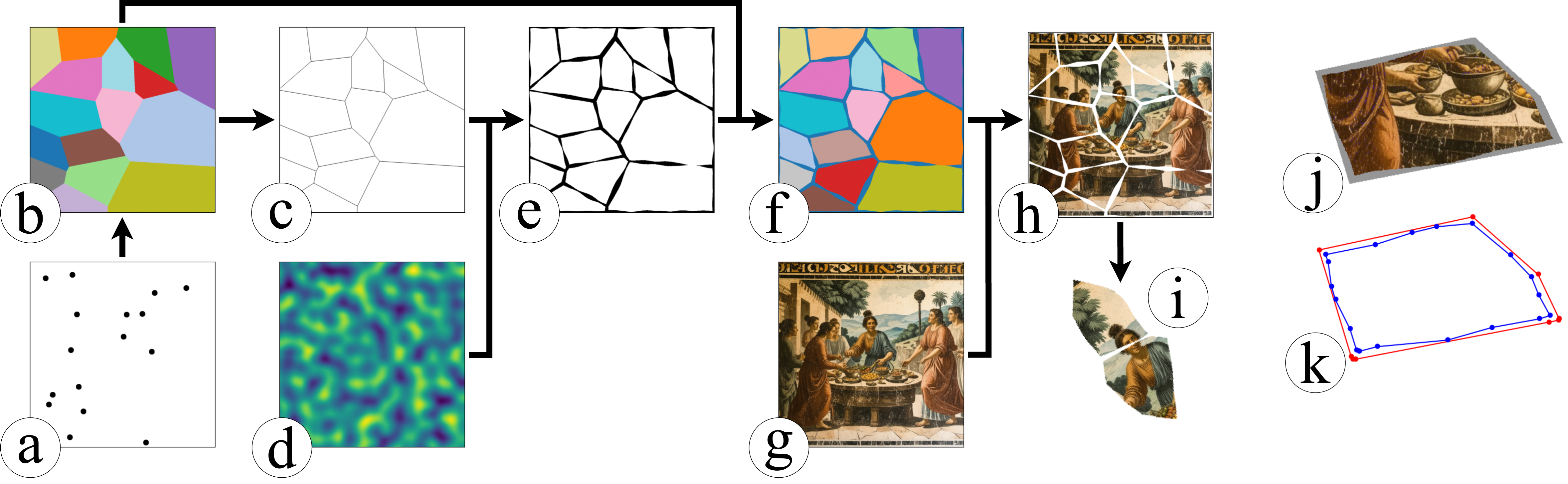}
    \caption{(a-h) Puzzle generation process. 
    (i) Pair extraction. 
    (j) An eroded fragment (pictorial) on top of its original Voronoi region (gray). 
    (k) Polygonization of the eroded fragment (blue) and Voronoi cell (red); note how the number of vertices increased post-erosion.
    }
    \label{fig:puzzle_gen}
\end{figure*}

\subsection{Puzzle and fragment pairs generation}

At the base of our dataset generation pipeline lies a Voronoi tessellation of the image domain, a mathematical process that partitions a plane into disjoint regions based on the distance to predetermined generating sites $\mathcal{V}(s_1,\dots,s_N)=\{R_1,\dots,R_N\}$, such that $R_i=\{x\in\mathbb{R}^2\mid\forall j,\Vert s_i-x\Vert\le\Vert s_j-x\Vert\}$. Thus, we start by uniformly sampling $s_1,\dots,s_N$ from the image plane (Fig.~\ref{fig:puzzle_gen}a) and treat each Voronoi region as a unique piece segment (Fig.~\ref{fig:puzzle_gen}b).

To simulate the natural erosion patterns observed in archaeological fragments, we implement a controlled degradation process on the Voronoi boundaries. First, we extract the region boundaries using an edge detector (Fig.~\ref{fig:puzzle_gen}c), creating a binary edge map that precisely localizes the potential erosion regions. We then generate a two-dimensional Perlin noise~\cite{perlin1985image} field matched to the image dimensions (Fig.~\ref{fig:puzzle_gen}d), which provides smoothly varying values that determine the local erosion intensity. This noise intensity at each edge pixel is then multiplied by the overall \textit{erosion rate} parameter, to determine an ``erosion radius'' around it. Pixels within this radius are then deleted from the Voronoi segmentation image (Fig.~\ref{fig:puzzle_gen}e-f). To produce the final puzzle, the eroded partition is applied to an image (Fig.~\ref{fig:puzzle_gen}h). 
Given the continuity of the noise model, this approach ensures spatial coherence in the erosion pattern while maintaining local variability, as adjacent boundary points with similar noise values undergo similar degrees of erosion. The resulting erosion map is then applied to the original Voronoi regions, discarding the eroded pixels. Finally, we use this eroded segmentation to extract the puzzle pieces from the input image, producing a set of realistically degraded fragments that closely mimic the appearance of actual archaeological artifacts.



To extract pairs of adjacent fragments from the generated puzzle (Fig.~\ref{fig:puzzle_gen}i), we relied on the Delaunay triangulation~\cite{delaunay1934sphere} of the generating sites $s_1,\dots,s_N$. This graph is known to be dual to the Voronoi diagram over the same set of site. Hence, fragments that are generated from adjacent sites in the Delauney triangulation are neighbors in the generated puzzle.

\subsection{Statistical properties of the proposed dataset}
\label{sec:statistics}

The puzzles created by the above Voronoi-based approach exhibit well-known statistical properties~\cite{meijering1953interface}.
Being generated from randomly distributed points, the expected region size in our Voronoi tessellation is $\frac{1}{N}$~\cite{meijering1953interface} of the total image area. This hints at the volume of pictorial information on a piece, intuitively implying that as the number of pieces increases, fragments that cover significant areas of the image are less likely. 
From a topological perspective, the tessellation forms a planar graph whose edges represent the interfaces between adjacent fragments. Following Euler's characteristic formula for planar graphs, the expected number of edges per fragment converges to 6 as $N$ increases~\cite{meijering1953interface}, which means each puzzle fragment is expected to have around 6 potential neighbors. Consequently,  the expected perimeter of each fragment is approximately $\frac{4}{\sqrt{N}}$~\cite{meijering1953interface}. Because compatibility often uses pictorial information at or near the boundary of fragments, this last figure indicates how much information should be analyzed during the reconstruction process.

\begin{figure}[ht]
    \centering
    \begin{tabular}{ccc}
        \includegraphics[width=0.31\linewidth]{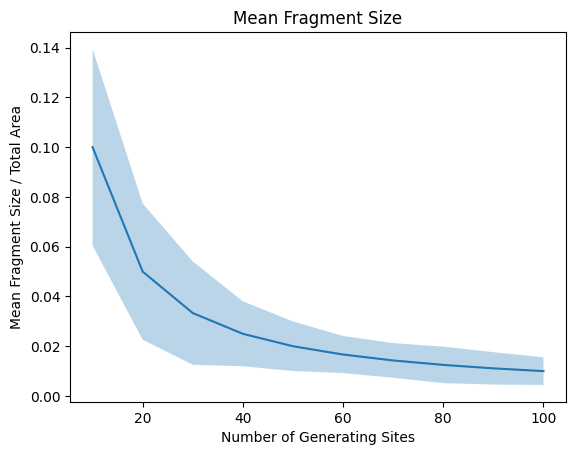} &
        \includegraphics[width=0.31\linewidth]{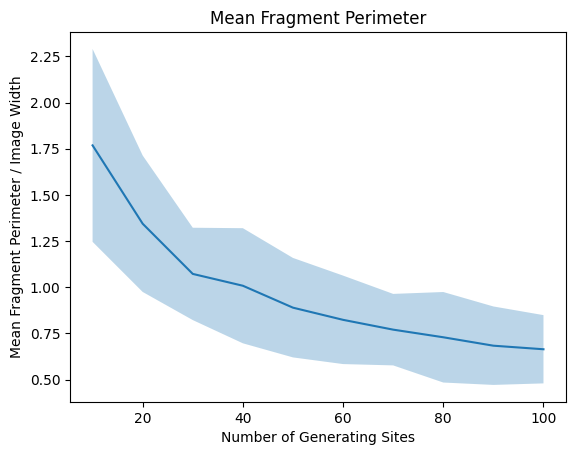} &
        \includegraphics[width=0.31\linewidth]{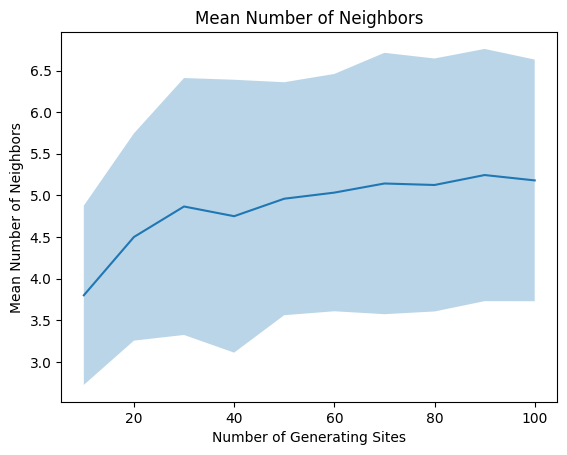} \\
        (a) & (b) & (c)
    \end{tabular}
    \caption{Empirical properties obtained from multiple non-eroded puzzles with 10, 20, \dots, 100 fragments.}
    \label{fig:puzzle_stats}
\end{figure}

As expected, certain properties, such as the area and perimeter of fragments, diverge from theoretical predictions as the erosion increases. While the neighboring relationships generally remain intact, the quantitative metrics should be interpreted as baseline expectations rather than precise values for heavily eroded puzzles. Figure~\ref{fig:puzzle_stats} presents empirical results to that effect.

\section{Pairwise alignment and compatibility for fragments of unrestricted shapes}
\label{sec:compatibility}
As briefly described above, the task of pairwise alignment, i.e., computing a valid alignment configuration out of an infinite space of potential transformations, is challenging. 
Nonetheless, the task of measuring compatibility, either geometric or pictorial in the presence of potential erosion, is challenging as well. 
With both in mind, PolEx is a compatibility measure that is completely free of restrictions regarding the geometric or pictorial features of the inspected fragments, or any information regarding erosion which might have altered them. 
To do so we take an approach which first reduces the otherwise infinite set of possible alignment configurations by utilizing the geometric shape of the fragments, and then scores them based on pictorial regions extrapolated beyond their original boundaries. 

Before we begin, we note that given the task of aligning two fragments, it is common to fix one while computing the proper alignment transformation for the other. In the following subsections we denote the latter the \textit{Source}, and the former the \textit{Target}.

\begin{figure}[]
    \centering
        \includegraphics[width=0.8\textwidth]{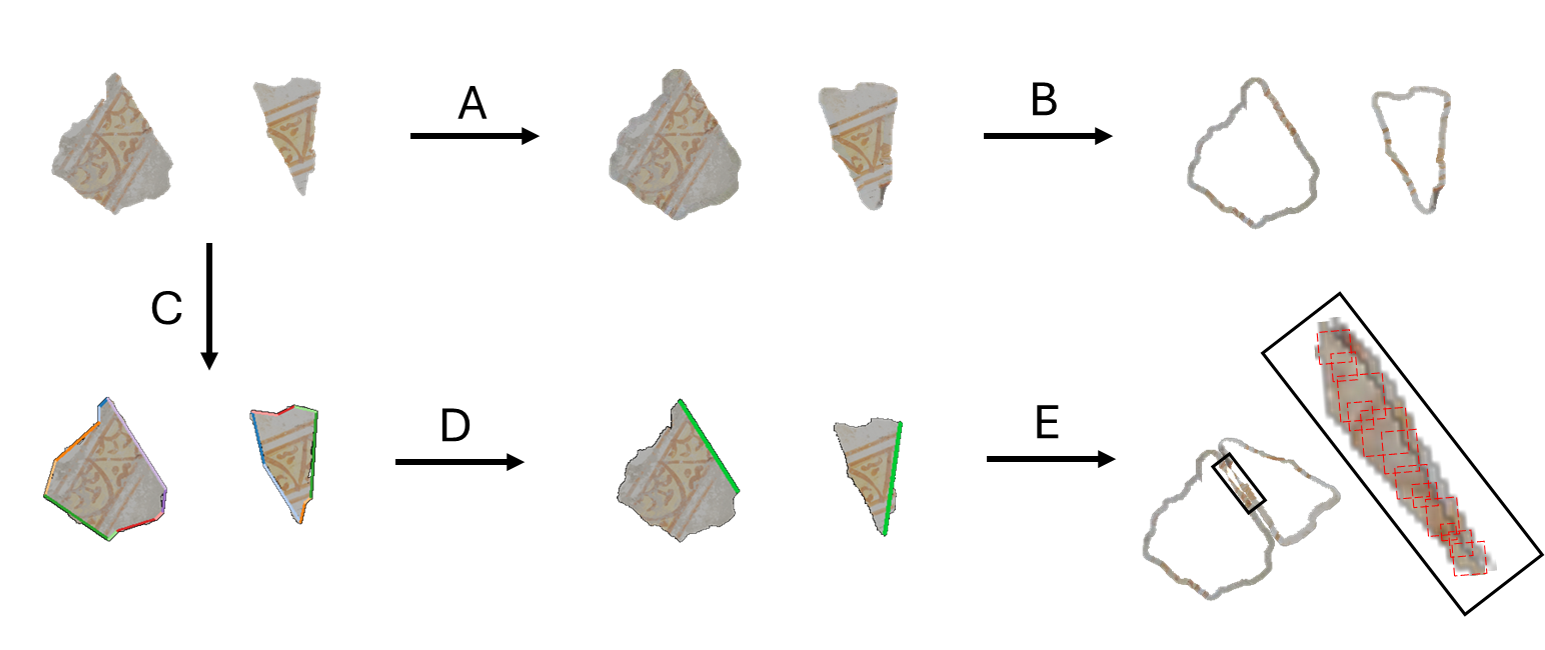}
    \caption{
            Illustration of the compatibility calculation process on two fragments from the RePAIR Dataset \cite{Tsesmelis2024}.
            \textbf{A:} Fragment extrapolations.
            \textbf{B:} Extraction of the extrapolated bands.
            \textbf{C:} Computation of the augmented polygonal approximations (colored edges).
            \textbf{D:} If a matched pair of edges (in green) is similar enough in length, an alignment transformation is computed. Note that the edge on the right fragment is an augmented edge.
            \textbf{E:} The alignment transformation is applied on the source fragment, and the shared region of the extrapolated bands is scored via aggregation of similarities of random patch pairs. 
            }
            \label{fig:repair_pieces_compatibility}
\end{figure}

\subsection{Handling erosion by extrapolation}
\label{sec:extrapolation}

Since fragments are eroded and thus miss regions along their perimeter, we first dilate their geometric shape and \textit{extrapolate} their pictorial content into the dilated area. This is done far enough to exceed their original (pre-eroded) boundary, while potentially utilizing some knowledge on the maximum erosion level. This is done to compensate for the missing near-border pictorial content that is typically used to evaluate the coherence of neighboring fragments.
The extrapolation is performed using a pre-trained Stable-Diffusion 1.4 model~\cite{Rombach2022}, which generates plausible visual extensions of the fragment. It should be noted that this pre-trained model was chosen as it was trained for the \textit{inpainting} task, for which extrapolation is a special (albeit extreme and more challenging) case. Hence, with proper adjustments, it could be applied on our task.  Moreover, this model can be activated without input text prompts, which greatly simplifies its application in our context. 

\begin{figure}
    \centering
    \includegraphics[width=0.8\linewidth]{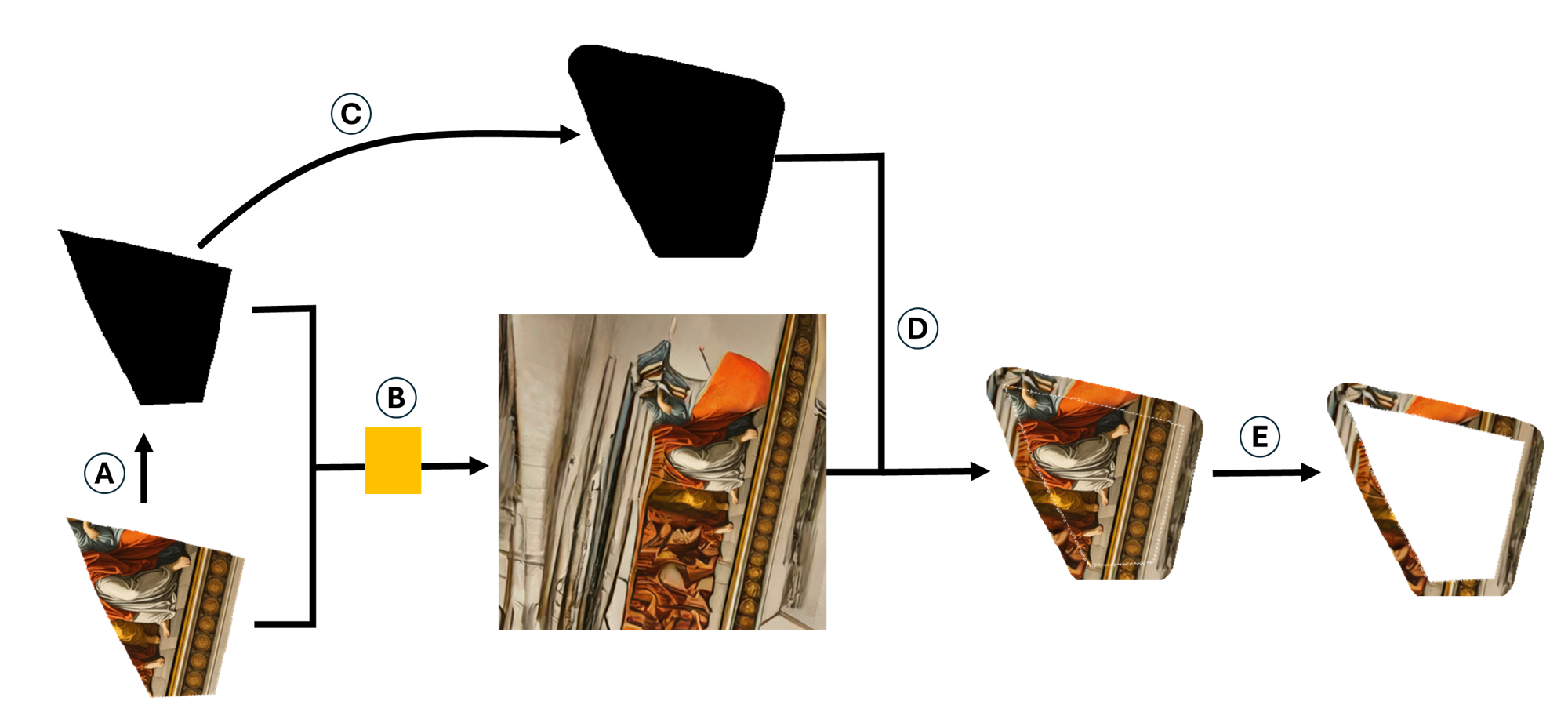}
    \caption{Fragment Extrapolation Process. First, the fragment's binary mask ($M$) is extracted (A), then both image and mask are fed into a Latent Diffusion Model~\cite{Rombach2022} (B) while $M$ is dilated into $M'$ (C). The model's output is then cropped by $M'$ to create an extrapolated fragment (D), from which the original content is removed to create the extrapolated band (E).}
    \label{fig:extrapolation}
\end{figure}

The extrapolation process extrapolates the fragment image with pictorial content to occupy the entire rectangular region of the image. To focus only on the relevant extrapolation regions, we apply a binary \textit{mask} that confines the newly generated content to a region slightly larger than the fragment’s original shape. 
Slightly more formally, let $I$ be the image of the fragment, and let $M$ be its binary mask (where $M=1$ indicates a fragment point and $M=0$ otherwise). 
Let 
$
M' = \text{dilate}(M, K_{\text{str}}),
$
be a morphologically dilated mask~\cite{serra1982}, where $K_{\text{str}}$is a standard structuring element of size $(n_{px} \times n_{px})$ (e.g., $n_{px}=20$).
$M$ is used as a "visual prompt" region for the Stable-Diffusion model, so that the generative process augments only the zone surrounding the fragment. 
The result is cropped back by $M'$, and the original content of the eroded piece is removed, yielding an \emph{extrapolated band} that possesses the extrapolated content around the original eroded fragment (see Fig.~\ref{fig:extrapolation} for an example).

\subsection{Space of potential alignments}

The space of potential relative alignments between two fragments is three-dimensional, continuous, and thus infinite. In contrast to earlier proposals, which sample this space uniformly and regularly (i.e., equally spaced samples) to allow reasonable exploration time~\cite{derech2021solving, Khoroshiltseva2024}, we seek an alternative approach that adapts the sampling to the geometrical shapes of the fragments. 
Specifically, we first compute explicit edges and augmented edges (see below) from the polygonal approximations of both fragments. Then, we compute the transformation required to align the selected edge of the source fragment to the one of the target, in order to score the pictorial coherency of the two fragments under the corresponding alignment.

\subsubsection{Polygonal approximation and edge augmentation}

In order to robustly capture the geometric shape of a fragment, we represent it as a binary image 
$
M(i,j)= 
\begin{cases}
1, & \text{if pixel } (i,j) \text{ belongs to the fragment}\cr
0, & \text{otherwise}
\end{cases}.
$


To obtain a clean boundary representation of the fragment, we apply a smoothing operation (e.g., Gaussian filtering) to $M$ to reduce noise and irregularities. We then re-binarize the result to extract a refined contour $C$ that is then is approximated by a polygon $P_C$ using the Ramer--Douglas--Peucker algorithm~\cite{douglas1973}. 

Let $p$ denote the perimeter of $P_C$ and let \mbox{$\varepsilon = \alpha \, p,$} with $\alpha \ll 1$ be a tolerance parameter. A representation simplification step
then retains only the vertices essential to the overall structure by removing the middle vertex where three consecutive vertices $\mathbf{p}_i$, $\mathbf{p}_{i+1}$, and $\mathbf{p}_{i+2}$ form an angle
$
\theta = \arccos\!\left(\frac{(\mathbf{p}_{i+1} - \mathbf{p}_i) \cdot (\mathbf{p}_{i+2} - \mathbf{p}_{i+1})}{\|\mathbf{p}_{i+1} - \mathbf{p}_i\|\,\|\mathbf{p}_{i+2} - \mathbf{p}_{i+1}\|}\right)
$
that is below a pre-defined small threshold $\Delta\alpha$. This process yields a compact polygonal representation of the fragment's shape (see Fig.~\ref{fig:repair_pieces_compatibility}).

To better capture geometric structure at different scales, and possible differences in polygonization of matching fragments, we augment the initial representation with additional edges. While there are different ways to augment the geometric representation, we chose to consider triples of consecutive \emph{edges} and test whether they can be considered one edge at some larger scale. If the central edge in the triple is sufficiently short and the outer edges are nearly collinear (i.e., their angular difference is below a given threshold), a new edge is synthesized from the first vertex of the first edge to the last vertex of the third edge. 

This augmented edge represents a larger-scale, implicit boundary segment that captures the overall geometric structure. In contrast to the edge-merging procedure performed in the prior polygonal simplification step, the augmented edge is \textit{added} to the representation as an extra edge, \textit{without replacing its components}. 
The augmented representation is no longer a simple polygonization, but a multiscale representation that captures numerous polygonizations concurrently.
This procedure is illustrated in Fig~\ref{fig:repair_pieces_compatibility}.

\subsubsection{Extraction of candidate matching edges}

To establish potential correspondences between the two fragments, candidate pairs of matching edges from the two fragments are identified from their polygonal approximations. 
Recall that each polygon is initially decomposed into base edges defined by consecutive vertices of the polygonization. 

Once the augmented representation is available for the source and target fragments, candidate matching between them is established based on edge lengths. Two candidate edges with lengths $L_t$ and $L_s$ are considered compatible if
$
\frac{\min(L_t, L_s)}{\max(L_t, L_s)} \geq \gamma,
$
where $\gamma$ is a predefined length ratio. 
Edge pairs failing to meet this criterion are considered inadmissible and discarded, an operation that filters out most candidate pairs for most practical values of $\gamma$ (see experimental evaluation). The pairs that survive this test proceed to the next step. 

\subsubsection{Calculating potential alignment configurations}

Given a pair of admissible candidate matching edges, our goal is to compute a rigid transformation that aligns the source edge with the target edge so that they are oppositely oriented and offset properly. Let $M_t$ and $M_s$ be the midpoints of the target and source edges, respectively. To ensure a proper gap between the fragments such that the erosion is accounted for, the target midpoint is offset outward by a distance $g$ along its outward normal $\mathbf{n}_t$, yielding an adjusted midpoint
$
M_t^{\text{offset}} = M_t + g\, \mathbf{n}_t.
$
Let the target edge have an orientation $\theta_t$ and the source edge an orientation $\theta_s$. The angle
$ \theta = (\theta_t + \pi) - \theta_s $
thus defines a rotation that aligns the two fragments such that the source edge faces its corresponding target edge, as would be expected if the two fragments indeed match this way. The source edge midpoint is then rotated by $\theta$ about the source fragment centroid to obtain $M_s^{\text{rot}}$. The required translation $T = (T_x, T_y)$ is computed as
$ T = M_t^{\text{offset}} - M_s^{\text{rot}} $
and the overall alignment configuration for this edge pair is thus given by the transformation parameters $(\theta, T_x, T_y)$, which are subsequently used to assess the pictorial compatibility of the fragment pair.

\subsection{Pictorial compatibility of candidate alignments}

Each candidate alignment is finally assessed for its pictorial compatibility by comparing the extrapolated bands of the fragments after the source fragment is transformed by $(\theta, T_x, T_y)$.
To do so we find the intersection of the two extrapolated bands and sample patches of random sizes in the predetermined range $[P_{\min}, P_{\max}]$ at fixed intervals with stride $s$. Given corresponding patches from the two bands, we then compute their LAB color difference. 
Since the latter is sensitive to the size of patch in which it is computed (i.e., indeed, small objects may be washed out in excessively large patches, and large objects may be missed in patches that are smaller), fixed size patches are intrinsically inferior. The random patches thus serve the purpose of better capturing pictorial structure, shapes, and details of unpredictable scales. 

More formally, for each patch $i$, let $\Omega_i$ denote its set of pixel indices. For each pixel $k\in\Omega_i$, let $(L_{t,k}, a_{t,k}, b_{t,k})$ and $(L_{s,k}, a_{s,k}, b_{s,k})$ denote the LAB color values in the target and source patches, respectively. The per-patch dissimilarity is then computed as
$
\Delta E_i = \frac{1}{|\Omega_i|}\sum_{k\in\Omega_i}\Delta E_{i,k},
$
where
$
\Delta E_{i,k} = \sqrt{(L_{t,k} - L_{s,k})^2 + (a_{t,k} - a_{s,k})^2 + (b_{t,k} - b_{s,k})^2}\,.
$

Since different pairs of patches yield different differences, all of the latter are aggregated using a $p$-norm over $N$ patches, namely
$
S = \left(\frac{1}{N}\sum_{i=1}^{N}\Delta E_i^p\right)^{1/p}\,,
$
and the value is finally normalized by the maximum possible LAB distance to yield a unitless measure. 

In order to handle extreme cases in which a large shared extrapolated region between the two fragments is similar and uniform (as in regions containing mostly sky or water) except for a small sub region, it is desirable to enhance the weight of such non-coherent sub regions in the computation. For that, we define an additional amplification factor $\lambda \geq 1$ that scales the score if prominent color exceptions are detected in the shared areas of the source and target extrapolations. 



\section{Evaluation}
\label{sec:evaluation}

As thoroughly described in previous works \cite{paumard2020deepzzle, Harel2024, Tsesmelis2024}, evaluating the reconstruction of fragment assemblies is challenging, not only because the evaluation metric must be invariant to rigid transformations, but because erosion introduces uncertainties in the desired outcome. 
At the same time, the evaluation measures previously proposed emerge from the need to score \textit{assemblies of many fragments}, while in our case we seek to handle just pairs.  With this in mind, we propose two pairwise metrics inspired by and improve upon the global metrics introduced by Tsemelis~\etal\cite{Tsesmelis2024} -- the \textit{Pairwise Relative Position Score} and the \textit{Pairwise Anchored RMSE} (of both the translation $T$ and rotation angle $rot$) -- to assess the quality of the predicted configurations relative to the ground truth.

For the \textbf{Pairwise Anchored RMSE}, the target fragment (used as the anchor) is first aligned according to its ground truth transformation. The source fragment is then transformed using its predicted configuration relative to the aligned target fragment. Denoting by $\mathbf{c}_p$ and $\mathbf{c}_g$ the centroids of the source fragment in the predicted and ground truth configurations respectively, the \textit{translation error} is defined as:
$
\text{RMSE}_{T} = \sqrt{(c_{p,x} - c_{g,x})^2 + (c_{p,y} - c_{g,y})^2}\,.
$
Similarly, let $\theta_p$ and $\theta_g$ denote the predicted and ground‐truth rotation angles of the source fragment. The \textbf{rotation error} is then defined as
$
\mathrm{RMSE}_{R}
= \min\!\bigl(\,|\theta_p - \theta_g|\,,\;2\pi - |\theta_p - \theta_g|\,\bigr)\,
$,
which also accounts for periodicity.

The \textbf{Pairwise Relative Position Score} quantifies the spatial consistency between the aligned configurations. After aligning the target and source fragments into a common coordinate frame, we extract the shared region (i.e., the intersection of their $M$ shapes as defined in Sec.~\ref{sec:compatibility}). The score is then defined as the ratio of the area of the shared region to the area of the ground truth fragment:
$
S_{\text{rel}} = \frac{\text{Area}(\text{Shared region})}{\text{Area}(\text{Source fragment})}\,.
$
A higher $S_{\text{rel}}$ score indicates better alignment between the fragments, and shorter adjustments are required to align the inspected solution with the ground truth solution. 
Together, the $S_{\text{rel}}$ and the RMSE metrics provide a comprehensive assessment of the geometric and spatial fidelity of the reconstruction.

\section{Results}
\label{sec:results}


\begin{figure}
    \centering
    \includegraphics[width=0.80\textwidth]{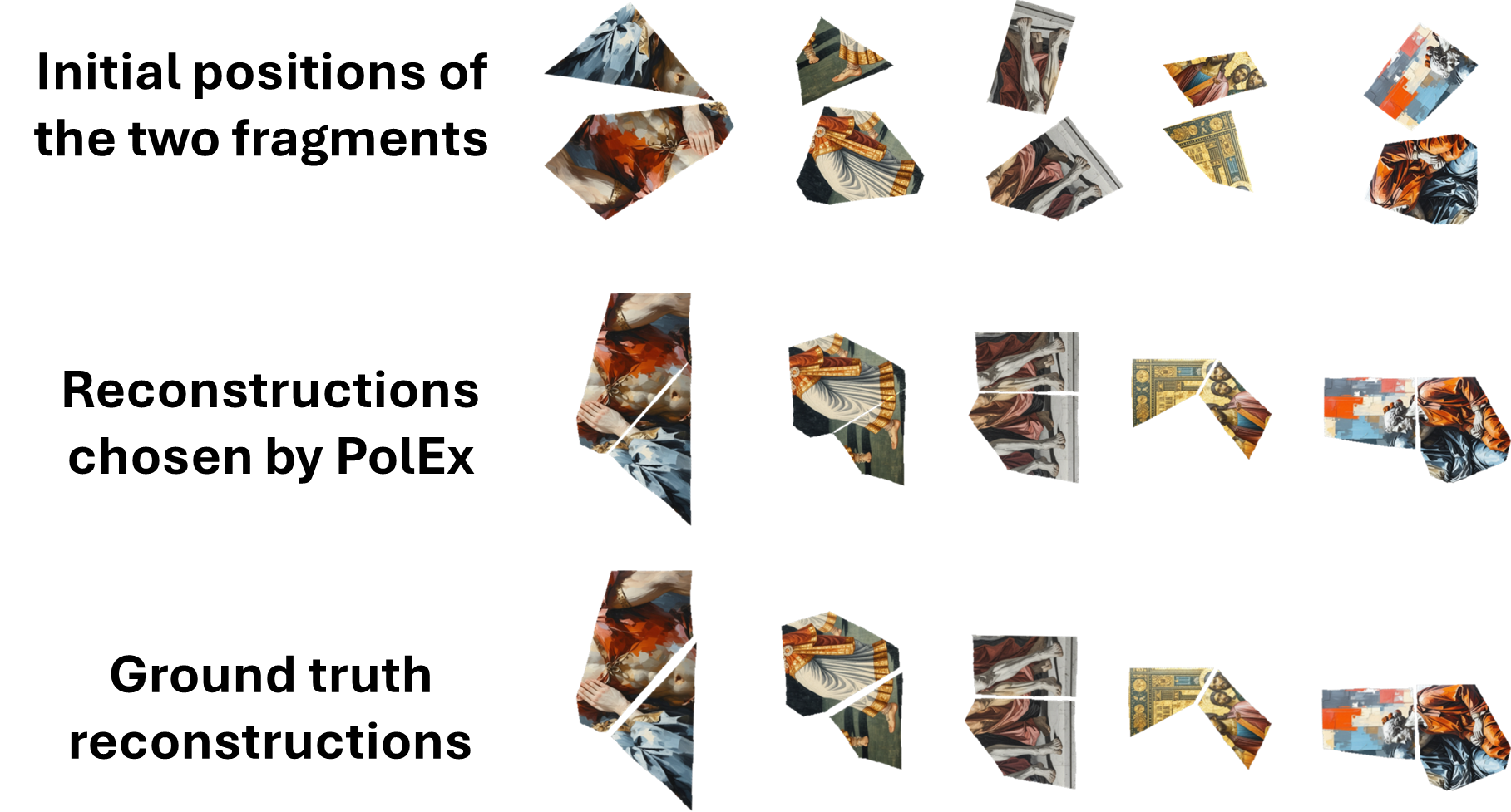}

    \caption{
            Qualitative Reconstruction Results on pairs of fragments from our dataset.
            }
            \label{fig:qualitative_solver_results}
\end{figure}

\begin{figure}
    \centering
        \includegraphics[width=0.3\textwidth]{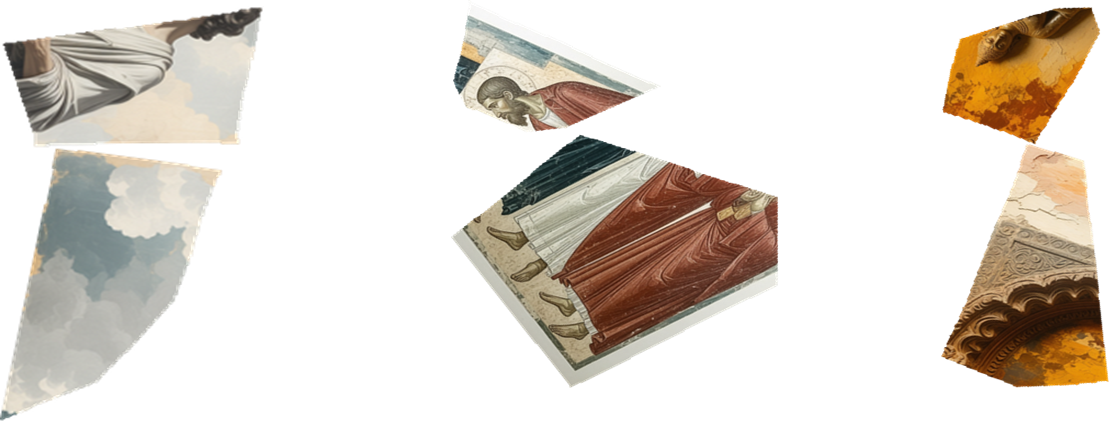}
        \\
        \includegraphics[width=0.3\textwidth]{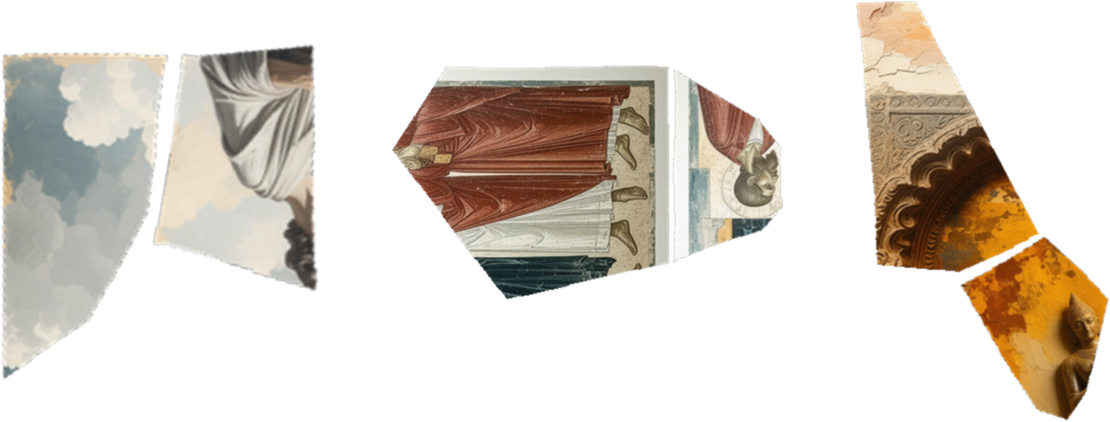}
        \\
        \includegraphics[width=0.3\textwidth]{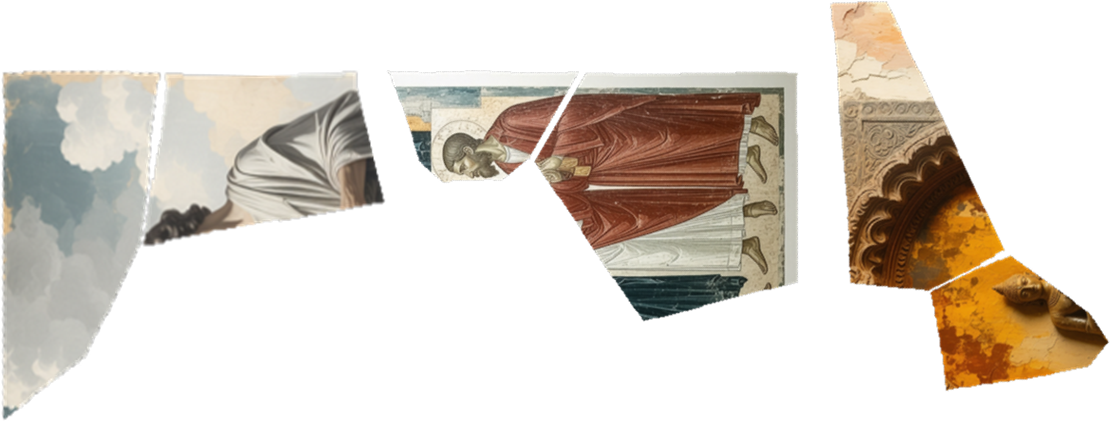}
        \\

    \caption{
            Potential pitfalls of our approach.
            \textbf{Top:} Initial positions of the two fragments.
            \textbf{Middle:} The reconstruction chosen by PolEx.
            \textbf{Bottom:} The ground truth reconstruction.
            }
            \label{fig:pitfalls}
\end{figure}

\begin{figure}
    \centering
    \includegraphics[width=1\textwidth]{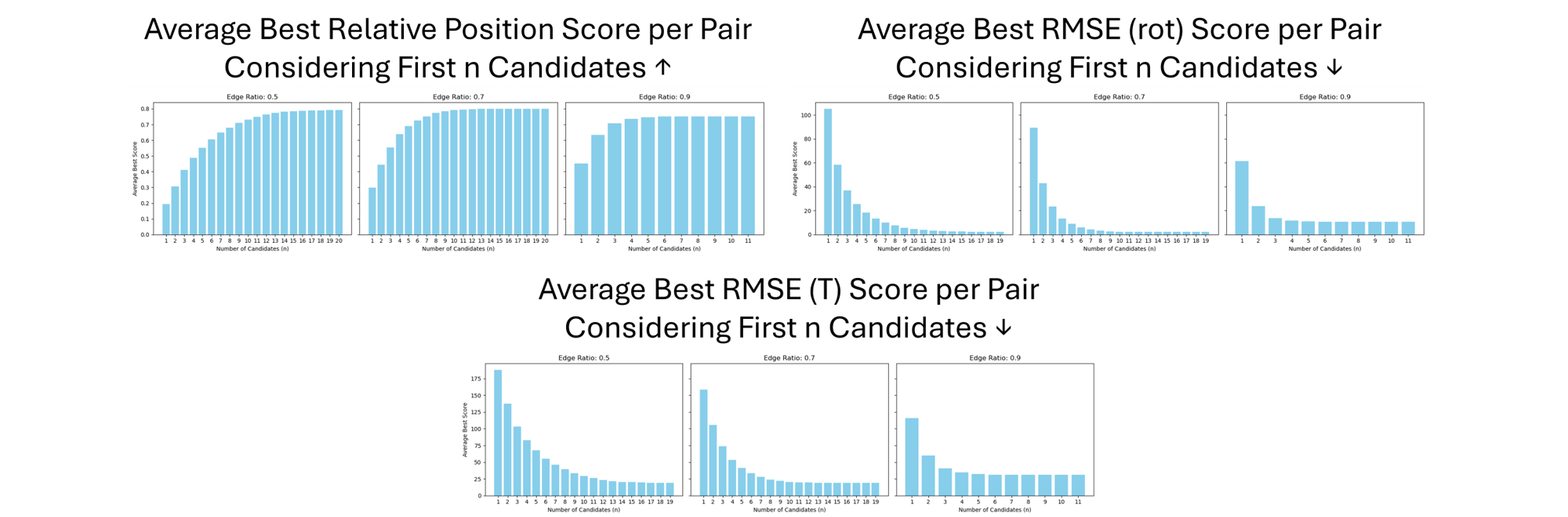}
    \\
    \caption{Quantitative results on all pairs in the test set, with different $\gamma$'s. Plots present the score averages (over all pairs, and each metric) of the best $n$ candidates. 
    }
    \label{fig:quantitative_results}
\end{figure}

In order to test the approach, we've applied PolEx on a test-set containing all of the pairs of neighboring fragments in 150 puzzles (with a total of 1483 fragments), randomly sampled from our dataset. We present comprehensive quantitative results in Fig. \ref{fig:quantitative_results}, using the three pairwise-adjusted evaluation metrics proposed in Sec. \ref{sec:evaluation}, while testing the edge matching procedure using different parameter values. 
As presented in Fig. \ref{fig:quantitative_results}, choosing only the top-ranked alignment configuration candidate already leads in many cases to the true one. However, a perfect compatibility cannot be expected due to the intractable nature of the puzzle-solving problem~\cite{demaine2007jigsaw}, and thus most solvers must assume a compatibility where the correct match is one among several best ones. In our case, considering multiple top-ranked candidates rapidly improves the overall scores in all measures. This can be explained by the fact that in some cases, our approach fails to rank the true reconstruction configuration of fragments as the top candidate, due to several potential confounds.
For example, cases in which fragments might share a relatively large and monochromatic region (such as a partial sky, or a part of a colored frame), may lead to preferring a transformation that aligns such region over the true shared region, especially if its smaller in size (See demonstration in Fig. \ref{fig:pitfalls}). 
Moreover, the potential change in the fragments' geometric shapes due to erosion, may too affect the possibility of matching truly neighboring edges, most commonly when an edge gets \textit{broken} into multiple non-collinear edges during polygonization (e.g., Fig.~\ref{fig:puzzle_gen}). 

In addition, while not our main goal, we integrated PolEx compatibility into a beam-search reconstruction framework (see Appendix A), which maintains a beam of top partial assemblies and iteratively extends each by the best-scored PolEx match before pruning to the top \(B\) hypotheses. On the RePAIR 2D dataset~\cite{Tsesmelis2024}, this solver achieves state-of-the-art results on all neighbor-based metrics (neighborhood precision, recall, F1) and remains competitive on remaining metrics.

%

\begin{table}[]
\small
  \centering
  \caption{Results on RePAIR 2D Dataset test-set \cite{Tsesmelis2024}}
  \label{tab:2d_results}
  \begin{threeparttable}
    \resizebox{\linewidth}{!}{%
      \begin{tabular}{lcccccc}
        \toprule
        Method & $Q_{pos}$ ↑ & RMSE ($\mathcal{R}^\circ$) ↓ & RMSE ($\mathcal{T}_{mm}$) ↓ & Precision ↑ & Recall ↑ & F1 ↑ \\
        \midrule
        Derech \etal\cite{derech2021solving} & 0.037 & \underline{80.964} & \underline{139.495} & \underline{0.454} & 0.527 & \underline{0.471}\\
        Tsemelis \etal~\cite{Tsesmelis2024} Genetic                & \textbf{0.047} & 85.625  & 151.714 & 0.313 & \underline{0.662} & 0.394\\
        Tsemelis \etal~\cite{Tsesmelis2024} Greedy                    & 0.023 & \textbf{76.987}  & \textbf{135.946} & 0.297 & 0.470 & 0.351\\
        Beam-search + PolEx                    & \underline{0.040} & 88.950  & 149.543 & \textbf{0.645} & \textbf{0.666} & \textbf{0.518}\\
        \bottomrule
      \end{tabular}
    }
  \end{threeparttable}
  \label{tab:repair_quantitive_results}
\end{table}

The experiments were conducted on a machine equipped with a 13th Gen Intel\textregistered\ Core\texttrademark\ i7-13620H processor (base clock 2.40 GHz), 32.0 GB of RAM, and an NVIDIA GeForce RTX 4070 Laptop GPU. The parameters used for evaluation were
$
\ell_{\min}=15\text{px},\quad \gamma=\tfrac{L_{\min}}{L_{\max}}=0.5,\quad
\Delta\alpha=10^\circ,\quad 
\alpha=0.005,\quad
k_{\mathrm{smooth}}=3,\quad 
g=10\text{px}
$. All values were selected after simple rough grid search on their ranges.

\section{Conclusions}

In this work we propose PolEx, an approach for determining pairwise compatibility of fragment images with unrestricted shapes, while computing optimal alignment configurations. We also provide a unique dataset and demonstrate results on it. As can be seen in Table~\ref{tab:repair_quantitive_results}, incorporating PolEx in an iterative puzzle-solving framework, achieves SOTA results on the RePAIR 2D dataset in all \textit{neighbor} or pairwise metrics presented in their work ~\cite{Tsesmelis2024}. Future works could consider incorporating it within a global-optimization based scheme, which could lead to better handling of potential pitfalls, as described in Sec.~\ref{sec:results}. 

\section*{Acknowledgments}
This work has been funded in part by the European Union’s Horizon 2020 research and innovation programme under grant agreement No 964854 (the RePAIR project). We also thank the Helmsley Charitable Trust through the ABC Robotics Initiative and the Frankel Fund of the Computer Science Department at Ben-Gurion University for their generous support.

\printbibliography[heading=bibintoc,title={References}]

\appendix
\section*{Appendix}

\section{PolEx compatibility in puzzle solving}

\label{sec:solver}

While this is not meant to be a puzzle-solving work, in order to demonstrate the effectiveness of PolEx, we embedded it within a novel archaeological reconstruction framework that generalizes the greedy strategies proposed in various prior works \cite{derech2021solving, Tsesmelis2024} by employing beam search. Reconstruction begins with a seed fragment, and during each iteration, every remaining piece is evaluated via our PolEx approach, which computes a composite score combining pictorial dissimilarity and an overlap penalty. 
Instead of a purely greedy approach, the framework maintains a beam of top hypotheses. Each hypothesis is extended by adding the fragment with the best score, and the beam is pruned to retain only the most promising candidates. This multi-path exploration reduces the risk of local minima and robustly assembles the fragments into a final reconstruction. The high-level pseudocode is shown in Algorithm~\ref{alg:beam-PolEx} and qualitative results on a small group of fragments sampled from RePAIR dataset \cite{Tsesmelis2024} are provided in Fig.~\ref{fig:qualitative_solver_results}. A detailed illustration of a single step (combining two fragments) is depicted in section 2.

\begin{algorithm}[H]
  \captionsetup{labelformat=empty}  
  \caption{Beam-Search Assembly with PolEx}
  \label{alg:beam-PolEx}

  \begin{algorithmic}              
    \State \textbf{Input:} Set $F$ of fragments and extrapolations, compatibility parameters, beam size $B$, etc.
    \State \textbf{Initialize:}
    \Statex \quad– Select target fragment $D$ as the seed.
    \Statex \quad– RHC $\leftarrow\{D\}$; remove $D$ from $F$; beam $\leftarrow\{D\}$.

    \While{$F \neq \emptyset$}
      \For{each hypothesis $H$ in beam}
        \For{each candidate fragment $S \in F$}
          \State Compute transform $T_S$ of $S$ w.r.t.\ $H$.
          \State Evaluate score (dissimilarity + overlap penalty).
        \EndFor
        \State Extend $H$ by adding the best-scoring $S^*$.
      \EndFor
      \State Prune beam to the top $B$ hypotheses.
      \State Remove the newly added fragment(s) from $F$.
    \EndWhile

    \State \textbf{Output:} Final reconstruction hypothesis and corresponding Euclidean transformations.
  \end{algorithmic}
\end{algorithm}

\begin{figure}[]
    \centering
        \includegraphics[width=0.8\columnwidth]{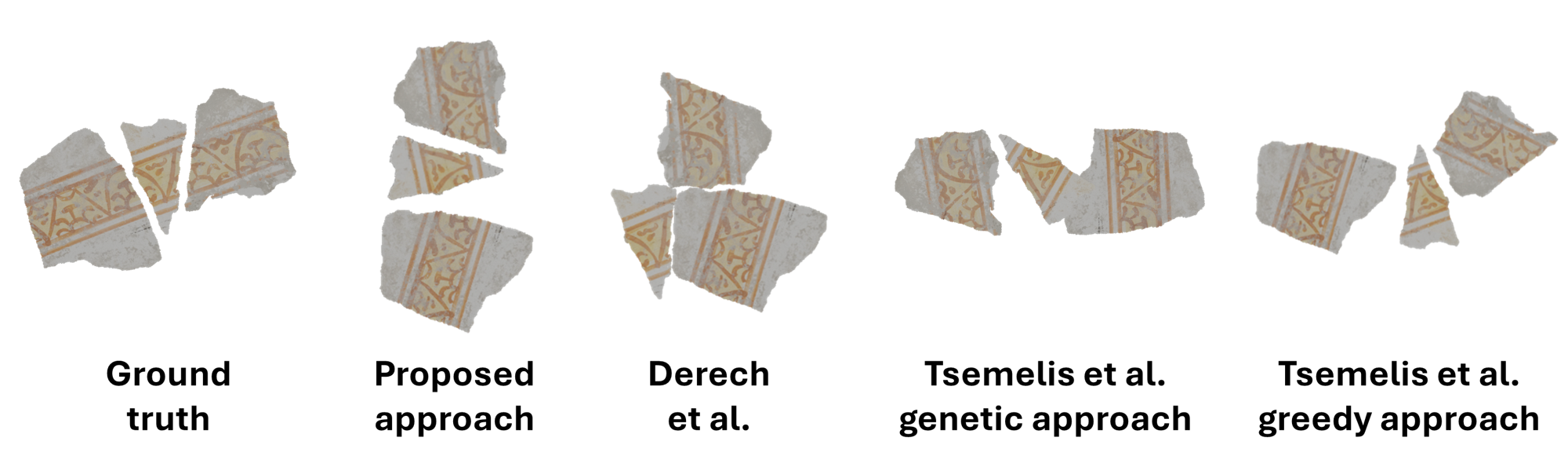}
    \caption{
            Qualitative reconstruction results on a selected fragments group from the RePAIR Dataset~\cite{Tsesmelis2024}.
            }
            \label{fig:qualitative_solver_results}
\end{figure}

\end{document}